# Marathi To English Neural Machine Translation With Near Perfect Corpus And Transformers

**Swapnil Ashok Jadhav,   Dailyhunt**

## Abstract

There have been very few attempts to benchmark performances of state-of-the-art algorithms for Neural Machine Translation task on Indian Languages. Google[1], Bing[2], Facebook and Yandex[3] are some of the very few companies which have built translation systems for few of the Indian Languages. Among them, translation results from Google are supposed to be better, based on general inspection. Bing-Translator do not even support Marathi language[4] which has around 95 million speakers and ranks $15^{th}$ in the world[5] in terms of combined primary and secondary speakers. In this exercise, we trained and compared variety of Neural Machine Marathi to English Translators trained with BERT-tokenizer by huggingface and various Transformer based architectures using Facebook's Fairseq platform with limited but almost correct parallel corpus to achieve better BLEU scores than Google on Tatoeba and Wikimedia open datasets.

## 1. Introduction & Related Work

In the last few years, neural machine translation (NMT) has achieved tremendous success in advancing the quality of machine translation (Cheng et al., 2016; Hieber et al., 2017). As an end-to-end sequence learning framework, NMT consists of two important components, the encoder and decoder, which are usually built on similar neural networks of different types, such as recurrent neural networks (Sutskever et al., 2014; Bahdanau et al., 2015; Chen et al., 2018), convolutional neural networks (Gehring et al., 2017), and more recently on transformer networks (Vaswani et al., 2017). To overcome the bottleneck of encoding the entire input sentence into a single vector, an attention mechanism was introduced, which further enhanced translation performance (Bahdanau et al., 2015). Deeper neural networks with increased model capacities in NMT have also been explored and shown promising results (Bapna et al., 2018).

Sequence-to-sequence neural models (seq2seq) (Kalchbrenner & Blunsom, 2013; Sutskever et al., 2014; Cho et al., 2014; Bahdanau et al., 2015) have been widely adopted as the state of-the-art approach for machine translation, both in the research community (Bojar et al., 2016; 2017; 2018) and for large-scale production systems (Cheng et al., 2016; Zhou et al., 2016; Crego et al., 2016; Hassan et al., 2018). As a highly expressive and abstract framework, seq2seq models can be trained to perform several tasks simultaneously (Luong et al., 2015), as exemplified by multilingual NMT (Dong et al., 2015; Firat et al., 2016; Ha et al., 2016; Johnson et al., 2017) - using a single model to translate between multiple languages.

Machine translation is highly sensitive to the quality and size of data. Mining sentence pairs using nearest neighbour with a hard threshold over cosine similarity method improves translation performance significantly as explained in (Artetxe & Schwenk, 2019). In our approach we did use cleaning of sentence pairs but we chose to go with simple dictionary based approach as explained in next section.

There are very few efforts where NMT with Indian languages are experimented and benchmarked properly. In recent approaches researchers have tried to build Multilingual NMT systems (Arivazhagan et al., 2019). Though, only few BLEU scores are presented for Indian languages and in most cases average BLEU score over multiple language-pairs has been considered as benchmark. In our approach explained in this paper, we have built a machine translation system with a widely spoken Indian language "Marathi" to English translation, with minimal engineering efforts for data collection & preparation and existing transformer architectures with easy-to-use Fairseq platform, which shows significant improvements compared to industry best Google Translator.

## 2. Data Collection & Preparation

We ideally need huge dataset of marathi-to-english parallel corpus to train deep architectures and to get state-of-the-art results with Fairseq platform. Opus[6] provides good amount of parallel corpus. For marathi-english pair, we can see that around 1 million sentences are available. Among which only Tatoeba, Wikimedia and bible datasets are useful, as other data is just instructions. Among the valid sets as well, when sanity check was done, it was found that not all sentences are correctly aligned and there some fetal mismatches. We tried to rectify but later decided not to and then ignored the bible dataset completely. We decided to keep Wikimedia and Tatoeba datasets for validation purpose as we were left with just around 53k sentence pairs and did not use the same in training.

---

[1] https://translate.google.com/
[2] https://www.bing.com/translator
[3] https://translate.yandex.com/
[4] https://en.wikipedia.org/wiki/Marathi_language
[5] https://en.wikipedia.org/wiki/List_of_languages_by_total_number_of_speakers

Through scrapping, we had more than 6 million sentence pairs, but we determined and used only those sentence pairs which were almost correct. We put a hard rule of dictionary-based words matching and considered only those sentence pairs which had at-least 30% of the translated words matched with dictionary words. We were left with around 3 million sentence pairs.

We used wordpiece tokenizer by huggingface[7] to tokenize Marathi and English text. Also, we used lowercased English text throughout the experiments to reduce the learning of cases for English language. There is no concept like "cases" for Marathi language. Sample parallel corpus examples can be found at project gihub location[8].

## 3. Experiments

### 3.1. Setup

We used Facebook's sequence-to-sequence library Fairseq[9] to train and inference the translation model. This neatly written and easy to use library provides multiple state-of-the-art architectures to build translation models. We installed the library on a 4x V100 32gb Nvidia GPU linux setup. Even though, there are multiple algorithms available, we focused majorly on following Transformer based architectures : *transformer-wmt-en-de, transformer-iwslt-de-en, transformer-wmt-en-de-big-t2t and transformer-vaswani-wmt-en-de-big*

Fairseq also provides option to tokenize the input text with sentencepiece tokenizer[10] and gpt tokenizer. But we opted to tokenize the text with wordpiece tokenizer instead, even before passing the text for the training. In future, we would try to build sentencepiece model from Marathi and English News corpus and use and evaluate against the existing wordpiece tokenizer.

### 3.2. Training

We trained multiple models with above mentioned transformer architectures with various hyper-parameters suggested in respective papers and in Fairseq github discussions. Following is the one of the training commands we used.

**Training Command :** *CUDA_VISIBLE_DEVICES=0,1,2,3 fairseq-train mr2en_token_data –arch transformer_vaswani_wmt_en_de_big –share-decoder-input-output-embed –optimizer adam –adam-betas '(0.9,0.98)' –clip-norm 0.0 –lr 5e-4 –lr-scheduler inverse_sqrt –warmup-updates 10000 –dropout 0.3 –weight-decay 0.0001 –criterion label_smoothed_cross_entropy –label-smoothing 0.1 –max-tokens 4096 –update-freq 2 –max-source-positions 512 –max-target-positions 512 –skip-invalid-size-inputs-valid-test*

Note that, as we were using 4 GPU's instead of 8 GPUs, mentioned in many state-of-the-art papers, we set *–update-freq* to 2. This is done to mimic the training with 8 GPUS. We used different optimizers but finally settled on *adam* optimizer because of its stable loss reduction capability. We noticed that increasing *warmup-updates* from 4k to 10k improved the convergence and reduced overall iterations. Also, note that we didn't use *–FP16* option in above command which could have improved the training speed but we observed that it reduces BLEU score marginally.

We stopped the training once perplexity(ppl) went below 3. Smaller models like *transformer-wmt-en-de* and *transformer-iwslt-de-en* took around 30 hours whereas, other two big models took 50+ hours. For all the transformer-based models, loss went below 1 for train and test sets.

### 3.3. Results

To make the comparison fair we used $16^{th}$ iteration of the models throughout. Marathi text was fired against the Google-cloud-api-v2 to collects the results for the comparison. Inference on GPU was preferred over CPUs as we could utilize all 4 GPUs effectively with 4 models. Following is the one of the command for inference we used.

**Inference Command :** *CUDA_VISIBLE_DEVICES=0 python interactive.py –path ../translation_task/checkpoints_transformer_iwslt_de_en/checkpoint16.pt ../translation_task/mr2en_token_data –beam 5 –source-lang mr –target-lang en –input ../translation_task/set3_tokens.mr –sacrebleu –skip-invalid-size-inputs-valid-test –batch-size 32 –buffer-size 32*

We used beam search of 5 which worked better in BLEU score than any other option. After the inference, we got tokenized English text which we de-tokenized and used further for model metrics comparisons. Note that, calculating BLEU with tokenized text yields high scores which is unfare

---

[6] http://opus.nlpl.eu/
[7] https://pypi.org/project/pytorch-pretrained-bert/
[8] https://github.com/swapniljadhav1921/marathi-2-english-NMT
[9] https://fairseq.readthedocs.io/
[10] https://github.com/google/sentencepiece

*Table 1.* BLEU score comparison on small sentences having word-count less than 15.

| Models | bleu | raw-bleu |
| --- | --- | --- |
| Google | 55.10 | 46.59 |
| wmt-en-de | 65.23 | 65.26 |
| iwslt-de-en | 63.11 | 63.13 |
| wmt-en-de-big-t2t | **71.97** | **71.99** |
| vaswani-wmt-en-de-big | **72.37** | **72.40** |

*Table 2.* BLEU score comparison on medium to large sentences having word-count more than 15.

| Models | bleu | raw-bleu |
| --- | --- | --- |
| Google | 28.60 | 17.47 |
| wmt-en-de | 26.87 | 26.10 |
| iwslt-de-en | 26.06 | 25.28 |
| wmt-en-de-big-t2t | **29.50** | **28.73** |
| vaswani-wmt-en-de-big | 27.18 | 26.50 |

to Google Translator and hence avoided throughout.

We used sacreBLEU[11] library to calculate corpus-bleu score (with smoothing function enabled) and raw-corpus-bleu score (with smoothing function disabled). As mentioned before we used, Tatoeba and Wikimedia parallel corpus of around 53k sentence pairs as validation set. Tatoeba contains smaller everyday sentences and greetings, while Wikimedia has long scientific sentences.

From Table 1, we can see that for smaller sentences having wordcount less than 15, all transformer models crushed Google in BLEU and Raw-BLEU scores. Boundary of 15 words was chosen based on study[12] which states that in current generation average words used in a sentence is around 10-20. Also note that, there is a less gap between BLEU and Raw-BLEU scores for all transformer models compared to Google. This table shows that for smaller everyday sentences and greetings, our model outperformed Google easily. vaswani-big architecture and wmt-t2t architecture performed the best.

From Table 2, we can see that all the models including Google struggled to go beyond 30 BLEU score. Also, only wmt-t2t model was able to outscore Google in BLEU but at the same time Google struggled in Raw-BLEU score compared all other models. This shows that wmt-t2t model was able to translate longer, complex sentences with good score and did better than Google.

Table 3 shows the comparison between actual English text and predicted English text word-counts. Even though, this score doesn't signify quality of the translation but often used to check the sanity of the model. Here we can see that, all transformer-based models outperformed Google in both Mean-Absolute-Error(MAE) and Root-Mean-Squred-Error(RMSE). Also, wmt-t2t model performed the best again overall.

---

[11]https://github.com/mjpost/sacreBLEU
[12]https://techcomm.nz/Story?Action=View&Story_id=106

*Table 3.* Error between actual translation word-count and predicted translation word-count.

| Models | MAE | RMSE |
| --- | --- | --- |
| Google | 1.257 | 7.239 |
| wmt-en-de | 0.783 | 6.965 |
| iwslt-de-en | 0.816 | 6.964 |
| wmt-en-de-big-t2t | **0.691** | **6.879** |
| vaswani-wmt-en-de-big | 0.723 | 7.149 |

*Table 4.* Comparison between existing Translators

| Marathi Text | आयुष्य पतंगासारखं आहे . मांजा धरला तर वेगात उंच झेपावत नाही आणि सोडला तर कुठे जाईल त्याचा नेम नाही. |
| --- | --- |
| Actual Translation | Life is like a kite. If you keep holding the thread, it will not rise faster and if you loose it then not sure where it will land . |
| Our Model | life is like a kite . holding a cat does not accelerate high speeds and does not specify where it will go if left unattended . |
| Google | Life is like a kite. If you catch a cat, it does not jump high and it does not specify where you will go. |
| Facebook | Life is like a fall. If you hold a cat, you don't run high in speed and if you leave it, there is no name where it goes . |
| Yandex | The life is like The Moth . I held her naked buttocks in my hands as she rode me until she climaxed . |

### 3.4. Discussion

Any language base model require huge amount of data to train deep architectures. We saw that one of the best word-embedding model BERT[13] was trained on more than 100gb of textual data. Similarly, to train translation models and to make them learn how to generate sentence structures and even transliterate proper nouns instead of translation, we need large parallel corpus. And Google supposedly have a very large corpus. As per our knowledge, Google relies on scrapping and community help[14]. As Google has not released any data, we cannot further comment on the data quality but note that we just used 3 million sentence pairs to cross Google's BLEU score.

---

[13]https://github.com/google-research/bert
[14]https://translate.google.com/community#mr/en

*Table 5.* Comparison between Google and Our Best Transformer Model

| Marathi Text 1 | "ज्यात पिल्ले तयार होतात असा पक्षी वा काही प्रकारच्या माद्यांपासून उत्पन्न होणारा गोलक, कोकिळा आपले अंडे उबवण्यासाठी कावळ्याच्या घरट्यात ठेवते." |
|---|---|
| Google Output 1 (Few Months Ago) | Spherul |
| Google Output 1 (Current) | The spider, the spider, lays its eggs in the nest to hatch. |
| Our Model Output 1 | a bird that produces chicks or a sphere of females keeps the cuckoo in a raven ' s nest to hatch its eggs |
| Marathi Text 2 | "आज १८ जून आहे व आज म्युरिएलचा वाढदिवस आहे !" |
| Google Output 2 | Today is June 5th, and today is Muriel's birthday! |
| Our Model Output 2 | today is june 18th and today is muriel ' s birthday ! |

Let's take one of the examples to see how various Translators performed and why we chose to compare our results with Google. Check the Table 4 for the results.

Yandex translation gives us an extreme example here. Somehow it converted philosophical saying into a adult-story content. Probably, the training dataset is the culprit here. But we also did check up with many other examples and noticed that Marathi-to-English translation provided by Yandex is pretty bad. Also, it looks like, it doesn't have enough vocabulary coverage as well.

Note that, none of the Translators including ours was able to capture the word "मांजा" which means "thread". Even if it looks wrong, it should be noted that our-model uses tokenization and possibly Google and Facebook as well. "cat" in Marathi translates to "मांजर" and due to tokenization it is very much possible that the word "thread" was translated to "cat" by above models. As we did not have any previous benchmarks, we tested all available translators with 500+ examples and selected Google translator to be compared with our transformer based models.

We will check few more example and compare our-model and Google output. The same can be checked in Table-5. In first example, Google few months ago produced only one word, and when we checked recently, it produced more than one word but completely missed the translation for a word "cuckoo" and ignored first half of the sentence. Our model overall produced correct sense out of the Marathi sentence. We suspect that, Google's new translation is from translation-api-v3 which is in beta phase or possibly due to regular model updates.

In another simple example, Google wrongly translated Marathi number "18" to "5". Also note that along with Google, our model did not try ro translate but transliterated the word "Muriel" which is proper noun here in the example. Our model also doesn't drop punctuations required for readability. More examples can be found in our github repository[15].

## 4. Conclusion & Future Work

From the results and examples we can see that our transformer-based model was able to outperform Google Translation with limited but almost correct parallel corpus. Google will keep improving its models and datasets, but with easy-to-use architectures like Fairseq, it is possible to compete with current state-of-the-art in the future.

This work suggests that limited datasets are often enough in improving translation accuracy if they are nearly clean. Also, there is a need of unified validation set on which researchers can benchmark their models. We propose that openly available "Tatoeba + Wikimedia" datasets should be considered to baseline the benchmarks for Marathi-to-English translation task.

We did use wordpiece tokenizer in this experiment, but we are planning to use our own sentencepiece tokenizers trained on massive news corpus each for Marathi and English and then repeat the exercise again and compare the results. We hope to see gain in BLEU score with this new tokenization method.

In future, we are planning to support multiple Indian languages for English-Translation task and also multilingual translation support and hope that, it will help us further in other NLP tasks like, Named-Entity-Recognition from news text and detecting similar news across languages.

## References

Arivazhagan, N., Bapna, A., Firat, O., Lepikhin, D., Johnson, M., Krikun, M., Chen, M. X., Cao, Y., Foster, G., Cherry, C., Macherey, W., Chen, Z., and Wu, Y. Massively Multilingual Neural Machine Translation in the Wild: Findings and Challenges. *arXiv e-prints*, art. arXiv:1907.05019, Jul 2019.

Artetxe, M. and Schwenk, H. Margin-based parallel corpus

---

[15] https://github.com/swapniljadhav1921/marathi-2-english-NMT

mining with multilingual sentence embeddings. In *Proceedings of the 57th Annual Meeting of the Association for Computational Linguistics*, 2019.

Bahdanau, D., Cho, K., and Bengio, Y. Neural machine translation by jointly learning to align and translate. In *International Conference on Learning Representations.*, 2015.

Bapna, A., Chen, M. X., Firat, O., Cao, Y., and Wu, Y. Training deeper neural machine translation models with transparent attention. In *arXiv preprint, arXiv:1808.07561.*, 2018.

Bojar, O., Chatterjee, R., and Federmann, C. Findings of the 2016 conference on machine translation. In *ACL 2016 FIRST CONFERENCE ON MACHINE TRANSLATION (WMT16), pages 131–198. The Association for Computational Linguistics.*, 2016.

Bojar, O., Chatterjee, R., and Federmann, C. Findings of the 2017 conference on machine translation (wmt17). In *Proceedings of the Second Conference on Machine Translation, pages 169–214.*, 2017.

Bojar, O., Federmann, C., Fishel, M., Graham, Y., Haddow, B., Koehn, P., and Monz, C. Findings of the 2018 conference on machine translation (wmt18). In *Proceedings of the Third Conference on Machine Translation: Shared Task Papers, pages 272–303, Belgium, Brussels. Association for Computational Linguistics.*, 2018.

Chen, M. X., Firat, O., Bapna, A., MelvJohnson, Macherey, W., Foster, G., Jones, L., Parmar, N., Schuster, M., and Chen, Z. The best of both worlds: Combining recent advances neural machine translation. In *Association for Computational Linguistics.*, 2018.

Cheng, Y., Xu, W., He, Z., He, W., Wu, H., Sun, M., and Liu, Y. Semisupervised learning for neural machine translation. In *Association for Computational Linguistics*, 2016.

Cho, K., van Merrienboer, B., Gulcehre, C., Bahdanau, D., Bougares, F., Schwenk, H., and Bengio, Y. Learning phrase representations using rnn encoder–decoder for statistical machine translation. In *Proceedings of the 2014 Conference on Empirical Methods in Natural Language Processing (EMNLP), pages 1724–1734, Doha, Qatar. Association for Computational Linguistics.*, 2014.

Crego, J. M., Kim, J., and Klein, G. Systran's pure neural machine translation systems. In *CoRR, abs/1610.05540.*, 2016.

Dong, D., Wu, H., He, W., Yu, D., and Wang, H. Multi-task learning for multiple language translation. In *Proceedings of the 53rd Annual Meeting of the Association for Computational Linguistics and the 7th International Joint Conference on Natural Language Processing (Volume 1: Long Papers), volume 1, pages 1723–1732.*, 2015.

Firat, O., Cho, K., and Bengio, Y. Multi-way, multilingual neural machine translation with a shared attention mechanism. In *arXiv preprint arXiv:1601.01073.*, 2016.

Gehring, J., Auli, M., Grangier, D., Yarats, D., and Dauphin, Y. N. Convolutional sequence to sequence learning. In *International Conference on Machine Learning.*, 2017.

Ha, T.-L., Niehues, J., and Waibel, A. H. Toward multilingual neural machine translation with universal encoder and decoder. In *CoRR, abs/1611.04798.*, 2016.

Hassan, H., Aue, A., and Chen, C. Achieving human parity on automatic chinese to english news translation. In *arXiv preprint arXiv:1803.05567.*, 2018.

Hieber, F., Domhan, T., Denkowski, M., Vilar, D., Sokolov, A., Clifton, A., and Post, M. Sockeye: A toolkit for neural machine translation. In *arXiv preprint, arXiv:1712.05690*, 2017.

Johnson, M., Schuster, M., and Le, Q. V. Google's multilingual neural machine translation system: Enabling zero-shot translation. In *Transactions of the Association of Computational Linguistics, 5(1):339–351.*, 2017.

Kalchbrenner, N. and Blunsom, P. Recurrent continuous translation models. In *Proceedings of the 2013 Conference on Empirical Methods in Natural Language Processing, pages 1700–1709, Seattle, Washington, USA. Association for Computational Linguistics.*, 2013.

Luong, M.-T., Le, Q. V., Sutskever, I., Vinyals, O., and Kaiser, L. Multitask sequence to sequence learning. In *arXiv preprint arXiv:1511.06114.*, 2015.

Sutskever, I., Vinyals, O., and Le, Q. V. Sequence to sequence learning with neural networks. In *Advances Neural Information Processing Systems*, 2014.

Vaswani, A., Shazeer, N., Parmar, N., Uszkoreit, J., Jones, L., Gomez, A. N., Łukasz Kaiser, and Polosukhin, I. Attention is all you need. In *Advances Neural Information Processing Systems.*, 2017.

Zhou, J., Cao, Y., Wang, X., Li, P., and Xu, W. Deep recurrent models with fast-forward connections for neural machine translation. In *Transactions of the Association for Computational Linguistics, 4:371–383.*, 2016.